\documentclass[conference]{IEEEtran}

\usepackage{array}
\IEEEoverridecommandlockouts
\usepackage{cite}
\usepackage{amsmath,amssymb,amsfonts}
\usepackage{algorithmic}
\usepackage{graphicx}
\usepackage{textcomp}
\usepackage{xcolor}
\usepackage{placeins}
\usepackage{url}
\usepackage{enumitem}
\def\BibTeX{{\rm B\kern-.05em{\sc i\kern-.025em b}\kern-.08em
    T\kern-.1667em\lower.7ex\hbox{E}\kern-.125emX}}

\usepackage{pgfplots}
\usepackage{tikz}
\usetikzlibrary{shapes.geometric, arrows}
\usetikzlibrary{patterns}
\pgfplotsset{compat=1.18}

\usepackage{filecontents}

\usepackage{flushend}

\begin{document}

\title{A Segmented Robot Grasping Perception Neural Network for Edge AI\\
}

\author{\IEEEauthorblockN{Casper Bröcheler, Thomas Vroom, Derrick Timmermans, Alan van den Akker,\\ Guangzhi Tang, Charalampos S. Kouzinopoulos, Rico Möckel}
\IEEEauthorblockA{\textit{Department of Advanced Computing Sciences} \\
\textit{Maastricht University}\\
Maastricht, Netherlands \\
\{c.brocheler, t.vroom, dxy.timmermans, alan.vandenakker\}@student.maastrichtuniversity.nl\\ \{guangzhi.tang, charis.kouzinopoulos, rico.mockel\}@maastrichtuniversity.nl
}

}

\maketitle

\begin{abstract}
Robotic grasping, the ability of robots to reliably secure and manipulate objects of varying shapes, sizes and orientations, is a complex task that requires precise perception and control.
Deep neural networks have shown remarkable success in grasp synthesis by learning rich and abstract representations of objects. When deployed at the edge, these models can enable low-latency, low-power inference, making real-time grasping feasible in resource-constrained environments.
This work implements Heatmap-Guided Grasp Detection, an end-to-end framework for the detection of 6-Dof grasp poses, on the GAP9 RISC-V System-on-Chip. The model is optimised using hardware-aware techniques, including input dimensionality reduction, model partitioning, and quantisation.
Experimental evaluation on the GraspNet-1Billion benchmark validates the feasibility of fully on-chip  inference, highlighting the potential of low-power MCUs for real-time, autonomous manipulation.
\end{abstract}


\section{Introduction} \label{sec:introduction}
Object grasping synthesis is a fundamental challenge in robotics, underpinning applications such as automated warehouse operations, patient assistance in healthcare, and object sorting on assembly lines \cite{tai2016state}. 
While humans excel at grasping objects of various shapes and sizes with precision, replicating this ability in robotics remains challenging. Insufficient grip strength may result in objects slipping, while excessive force risks damaging fragile or valuable items \cite{1308798}.

A recent trend in Deep Learning (DL) research is  \textit{Edge AI}, which shifts computation from the cloud to resource-constrained devices at the network's edge, enabling low-latency inference and re-training on energy-efficient Microcontroller Units (MCUs) \cite{jlpea14020019} \cite{brockmann2024optimizing} \cite{papaioannou2023ultra}. However, deploying DL models on MCUs for real-time robotic grasping presents significant challenges due to strict constraints on memory, processing power, and energy availability. Unlike cloud or high-performance computing paradigms, Edge AI implementations must be highly optimised, balancing model size, execution efficiency, and power consumption, while maintaining high levels of grasping accuracy and robustness.

This work performs robotic grasping inference at the edge. 
To reduce memory and computational requirements, a series of hardware-aware optimisation techniques are applied. These include the reduction of the input resolution from the original RGB-Depth (RGBD) data, employing pipelined execution to partition computation across sub-models, and quantising trained weights into compact, low-precision formats.

The proposed implementation targets the GreenWaves Technologies GAP9, which features RISC-V cores alongside a dedicated AI accelerator.
We demonstrate that real-time robotic grasp perception is achievable on low-power edge platforms by deploying HGGD-MCU, a hardware-optimised variant of the Heatmap-Guided Grasp Detection (HGGD) architecture\cite{chen2023efficient}.
The key contributions of this work are as follows: a) a pipeline methodology to efficiently downscale and run a large 6-Dof grasping model on a low-power MCU such as the GAP9; b) the benchmarking of the proposed method against other state-of-the-art models using average precision (AP), demonstrating competitive performance; c) the evaluation of the model’s inference time when executed on the MCU.


\section{Background} \label{sec:background}

\subsection{6D Grasping Perception} \label{background:6d}
6D grasping perception involves the manipulation of objects in six degrees of freedom: three translational (x, y, z) and three rotational (roll, pitch, yaw).
Current state-of-the-art techniques use RGBD cameras and point cloud representations to infer grasp configurations straight from raw sensory input \cite{agrawal2023real} \cite{xie2024rethinking}.

The main challenge in 6D grasping perception is the generalisation from training data to real-world viability \cite{bauer2024challenges}. Due to the complexity of dynamic real-world environments, grasping models have to generalise their training data to varying object geometries and material properties, while dealing with inconsistent environments and possibly limited visibility. Consequently, 6D grasping models require substantial high-quality training data and increased computational power for data processing \cite{chen2023efficient} \cite{bauer2024challenges}.

\subsection{GAP9}
GAP9 is a low-power, low-latency processor designed for efficient signal processing across voice, image, and audio modalities. It is optimised for real-time execution of Neural Network (NN) workloads, achieving high performance with reduced power consumption by leveraging a cluster of highly specialised RISC-V cores capable of executing floating-point multiply-accumulate (FMA) operations. This architecture mirrors the functionality of Neural Processing Units (NPUs) found in modern CPUs and System-on-Chips (SoCs), enabling accelerated tensor computation through hardware-level parallelism.  

The SoC operates at an internal clock frequency of up to $370$MHz, with Dynamic Voltage and Frequency Scaling. 
The architecture includes a Fabric Controller (FC) core that orchestrates system-level operations and a cluster of nine additional RISC-V cores, each with identical specifications to the FC. These cores execute independently, allowing concurrent processing of NN layers. A dedicated accelerator core, NE16, further enhances NN inference by offloading convolutional computations.

The FC provides $1.5$MB interleaved and $64$kB non-interleaved L$2$ memory, a $2$MB non-volatile eMRAM and a $2$kB instruction cache. The cluster has access to a $128$kB shared data memory, and features a hierarchical instruction cache with $4$kB shared memory and $6$kB private memory distributed among the cores.



\section{Methods} \label{sec:methods}

\begin{figure*}[h]
    \centering
    \includegraphics[scale=0.37]{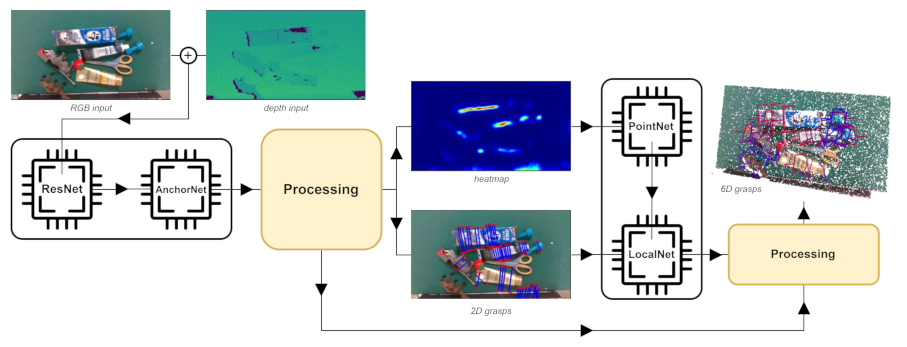}
    \caption{Edge grasping perception pipeline. The input is passed to a sequence of two MCUs running ResNet-MCU and AnchorNet-MCU. After post processing the output of AnchorNet-MCU, the resulting heatmaps and 2D grasps are used as input for the PointNet-MCU and LocalNet-MCU models running on two additional MCUs. The output of LocalNet-MCU together with  additional post processing can be subsequently converted into 6D grasps.}
    \label{fig:pipeline}
\end{figure*}

\subsection{Heatmap-Guided Grasp Detection (HGGD)}
For embedded grasping perception, we utilised the HGGD architecture. The architecture generates grasps by encoding grasp heatmaps from RGBD images, which guide the generation process. By processing only relevant image regions, it handles efficiently semantic and geometrical representations \cite{chen2023efficient}. HGGD consists of two separate models: AnchorNet to extract semantic features from the input and generates heatmaps, and LocalNet to generate grasps using a novel semantic-to-point feature extraction.

\subsubsection{AnchorNet}
An encoder-decoder model that generates heatmaps for different attributes of the RGBD image. The encoder is based on \textit{ResNet34}, a residual learning network with $34$ layers  \cite{he2016deep}. \textit{ResNet34} extracts features from the preprocessed RGBD image which are subsequently combined with various anchors (possible grasping points) sampled from the original image in the decoder \cite{zhou2018fully}. AnchorNet uses Convolutional Neural Networks (CNNs) to generate anchor-heatmaps for attributes such as width, rotation, and depth, identifying possible grasping regions.

\begin{figure}[h]
    \centering
    \includegraphics[scale=0.1]{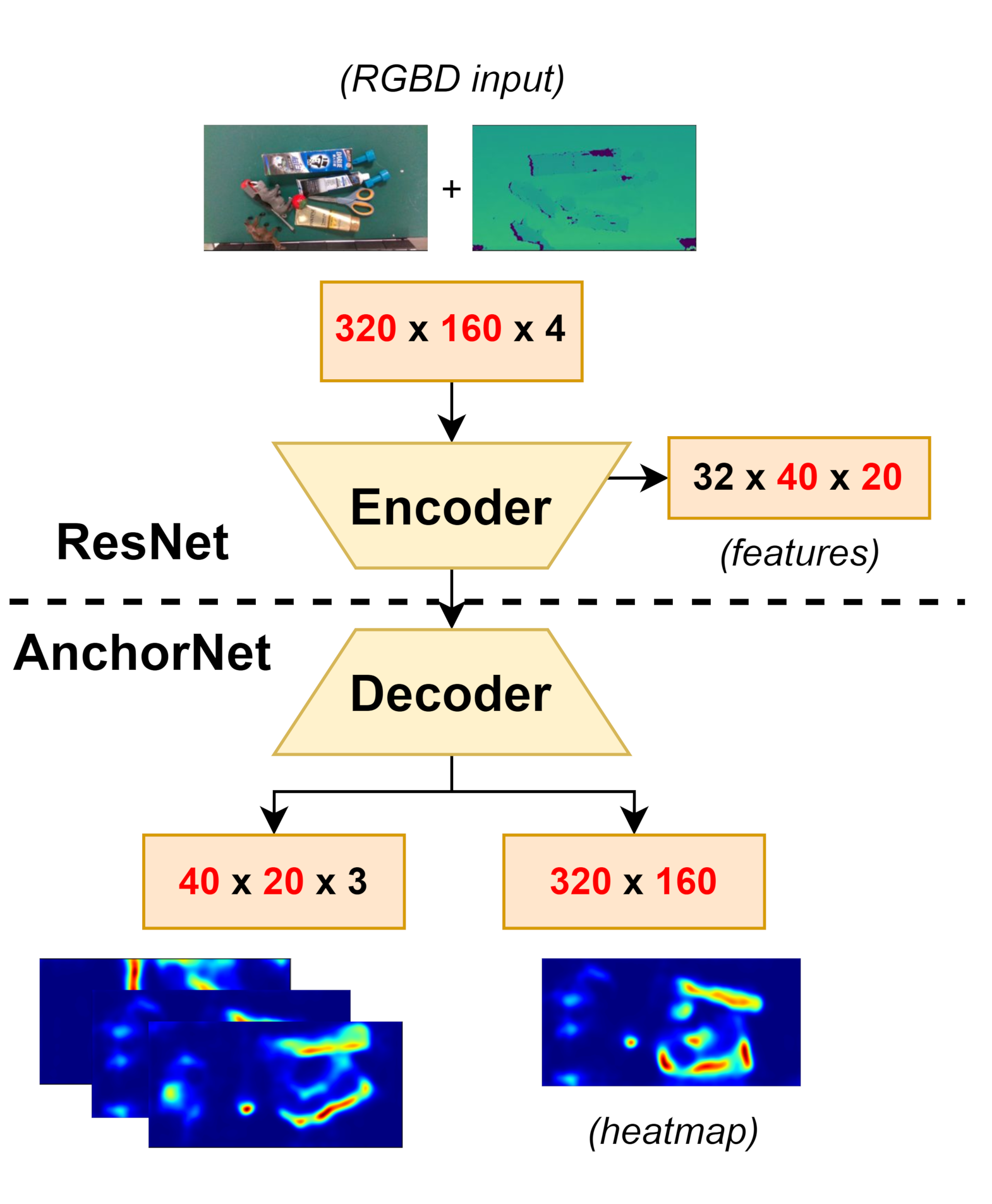}
    \caption{Adapted architecture of AnchorNet. The dotted line represents the partitioning point, where the original model is split into ResNet-MCU and AnchorNet-MCU.}
    \label{fig:anchornetmodified}
\end{figure}

\subsubsection{LocalNet}
LocalNet takes as an input a point cloud derived from the original RGBD image and a set of anchors sampled from AnchorNet. It then uses the grasp attributes of each anchor to predict the remaining attributes and create multiple valid 6D grasps.

Heatmap-identified regions likely to contain graspable objects, are mapped to the point cloud to locate the anchors in 3D space. The point cloud is subsequently combined with semantic features from AnchorNet, processed using a custom lightweight PointNet feature extractor, enabling a semantic-to-point feature extraction \cite{qi2017pointnet}. These features are used by a grasp generator to predict the missing attributes and refine the anchors from AnchorNet.

\begin{figure}[h]
    \centering
    \includegraphics[scale=0.08]{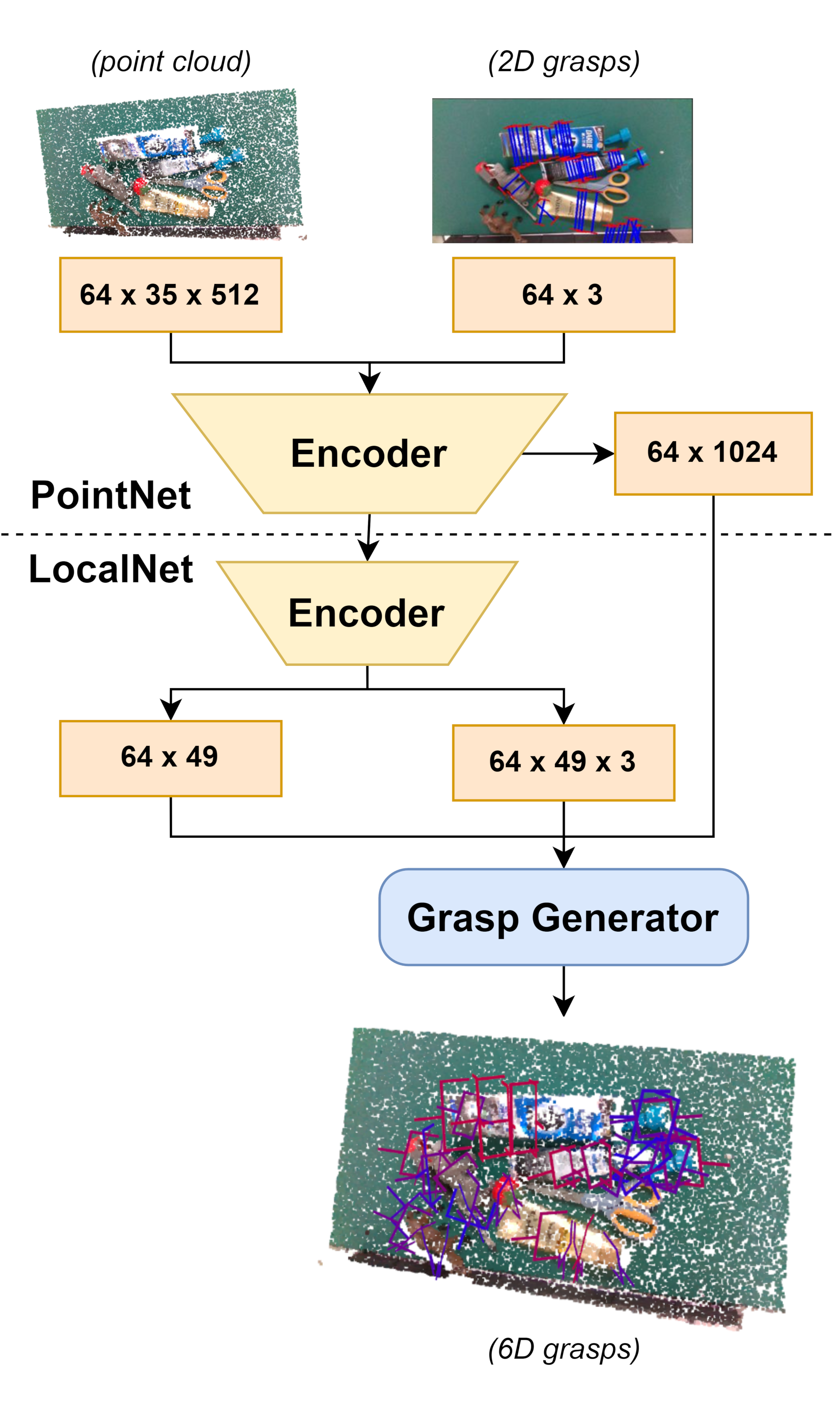}
    \caption{Adapted architecture of LocalNet. The dotted line represents the splitting point of the original model into PointNet-MCU and LocalNet-MCU.}
    \label{fig:localnetmodified}
\end{figure}

HGGD achieves state-of-the-art performance, with equivalent performance to models like GSNet over the GraspNet-1Billion dataset. It outperforms other models such as TransGrasp and REGNet \cite{liu2022transgrasp} \cite{zhao2021regnet}, especially in terms of inference speed, running more than twice as fast as GSNet \cite{wang2021graspness}.

\subsection{Optimisation Techniques} \label{sec:methods:optimisation}
This section discusses different optimisation techniques used to enable real-time MCU execution.

\subsubsection{Reducing Input Size}
One of the key optimisations for CNNs involves the reduction of the input image size \cite{wermter_2021_input}. Since CNNs store weights as kernels, lowering the image resolution can significantly reduce memory usage during convolution operations without impacting the model size. This is particularly important for the GAP9, which has limited L$2$ memory. By downscaling the input resolution from $640 \times 360$ to $320 \times 160$ pixels, the input size is reduced by approximately $75\%$, leading to improved memory usage and reduced latency due to faster transfers between L$2$ and L$3$ memory.

\subsubsection{Pipeline Execution}
To conserve RAM and flash, the model is partitioned into smaller functionally-equivalent sub-models, each processed sequentially. By utilising multiple MCUs in series, the processing of different sections of the network can be parallelised, with the output from one MCU being used as an input for the next. 
The model is partitioned into four sub-models, executed in the following order:

\begin{enumerate}
    \item ResNet-MCU (ResNet34 feature extractor / encoder)
    \item AnchorNet-MCU (heatmap generation / decoder)
    \item PointNet-MCU (semantic-to-point feature extraction)
    \item LocalNet-MCU (grasp refinement)
\end{enumerate}

The partitioning into the four sub-models allows a clear and consistent data flow, both computationally and semantically. Additionally, since none of the models are resized in any way, the weights from the original HGGD implementation can be losslessly transferred to the sub-models. The final adapted architecture of AnchorNet and LocalNet can be seen in Figures \ref{fig:anchornetmodified} and \ref{fig:localnetmodified} respectively.

\subsubsection{Quantisation}
Using the GAP9's scaled quantisation method \cite{dai2021vs}, the weights of the model were quantised from float32 to int8, reducing memory usage by a factor of four.  While quantisation may impact signal quality, prior studies show minimal practical effects on performance \cite{hashemi2017understanding}.

\subsubsection{Targeting Hardware Features}

To maximise inference performance on the GAP9, we leveraged the NE16 NN accelerator and parallel processing capabilities.
The NE16 is optimised for the height, width, channels (HWC) layout of input tensors, and the GAP9 clusters are designed for linear or piecewise data access. Efficient data management is critical, as the GAP9 does not incorporate any data caches. We used the Autotiler tool of the GAP9 SDK to optimise data layout and minimise data transfer latency.

\section{Experimental Methodology} \label{sec:experiments}
We conducted multiple experiments to evaluate our results, with detailed setup and methodology discussed in this section.

\subsection{Average Precision} \label{sec:EM AP}
To validate the proposed reduced pipeline, we replicated the original HGGD AP experiments and extended the evaluation by benchmarking against several state-of-the-art methods. The experiments mainly consisted of model evaluation on the testing data of the GraspNet-1Billion dataset \cite{fang2020graspnet}.

We recorded the AP of the reduced model on the Seen, Similar, and Novel subsets of the testing data, with friction coefficients of $\mu=1, 0.8, 0.4$. “Seen” data consists of objects present in the training data, “Similar” data includes unseen objects that look similar to ones seen in the training data, and “Novel” data consists of unseen objects that have little to no resemblance to objects seen in the training data.

To maintain consistency with the original HGGD study, we adopted the same set of hyperparameters. 
For a detailed discussion of their selection process and impact on model performance, the reader is referred to the original paper. The only additional hyperparameter introduced in this work is the input resolution, which was set to $320 \times 160$; the lowest resolution compatible with the original network.
At the same time, we incorporated the optimisation techniques discussed in section \ref{sec:methods:optimisation}, allowing for a direct comparison with both the GraspNet-1Billion baseline and the state-of-the-art GSNet.

The test set was executed on the GAP9 simulator GVSOC, and focused on the proposed pipeline execution approach, assessing its feasibility and performance on embedded hardware, without considering the data transfer overhead.


\subsection{Memory requirements} \label{sec:requirements exp}
To evaluate the impact of the different optimisation techniques on memory usage, we measured flash, RAM, and L$2$ memory consumption on the GAP9 platform.

A baseline was established using the partitioned sub-models with no optimisation applied to serve as a reference point. These are referred to as ``Original''. 




Different optimisation techniques were then applied to the decomposed models. More specifically, the input size of the data was reduced, the models were quantised from float32 to int8, and finally the number of model layers were reduced. The final, optimised models are labeled as ``Optimised''.



\subsection{Inference time} \label{sec:infernecetime exp}
Inference time was measured across all four sub-models. 
Execution times were recorded in processor cycles and later converted to milliseconds using the clock frequency of the MCU, based on the methodology of \cite{buturugua2017time}.

Each sub-model was executed $100$ times, and the resulting inference times were aggregated to estimate the total forward pass duration. Initial analysis revealed that the distribution of execution times was non-normal, preventing direct calculation of reliable confidence intervals. To address this, we employed bootstrap resampling: $25$ samples were randomly drawn with replacement from the measured inference times, repeated $400$ times to generate a bootstrapped distribution of the mean.
Normality of the bootstrapped means was verified using the Shapiro-Wilk test, which yielded a p-value of $0.265$, indicating no significant deviation from normality. This enabled the computation of statistically valid metrics, including t-values and confidence intervals.

\section{Results} \label{sec:results}

\subsection{Average Precision}

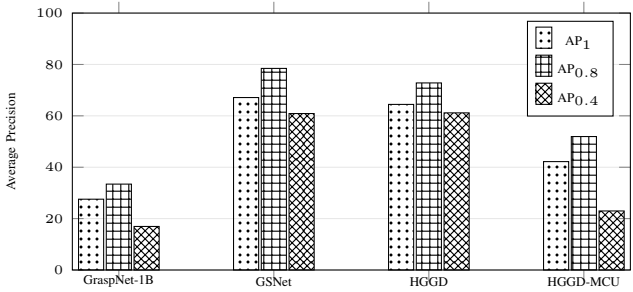
\begin{figure}[t]
	   \centering
		\begin{tikzpicture}[scale=1]
        \begin{axis}[
        ybar,
        bar width=9.5pt,
        width=9cm,
        height=5cm,
        grid=major,
        xmajorgrids=false,
        ymajorgrids=true,
        major grid style={line width=0.2pt, draw=gray!20},
        symbolic x coords={GraspNet-1B, GSNet, HGGD, HGGD-MCU},
        xtick=data,
        ymin=0, ymax=100,
        ybar=1pt,
        ylabel={\tiny Average Precision},
        xlabel={},
        tick label style={font=\tiny},
        ylabel near ticks,
        legend pos=north east,
        legend style={font=\tiny},
        x tick label style={rotate=0, anchor=center}
    ]
        \addplot[black!99, pattern=dots, single ybar legend] coordinates {(GraspNet-1B, 27.56) (GSNet, 67.12) (HGGD, 64.45) (HGGD-MCU, 42.19)};
        \addplot[black!99, pattern=grid,single ybar legend,] coordinates {(GraspNet-1B, 33.43) (GSNet, 78.46) (HGGD, 72.81) (HGGD-MCU, 51.94)};
        \addplot[black!99, pattern=crosshatch,single ybar legend,] coordinates {(GraspNet-1B, 16.95) (GSNet, 60.9) (HGGD, 61.16) (HGGD-MCU, 22.98)};
        \legend{AP$_1$, AP$_{0.8}$, AP$_{0.4}$}
    \end{axis}
\end{tikzpicture}
		\caption{Measured AP for the ``Seen'' dataset.}
		\label{fig:APseen}
	\end{figure}

\begin{figure}[t]
	   \centering
		\begin{tikzpicture}[scale=1]
        \begin{axis}[
        ybar,
        bar width=9.5pt,
        width=9cm,
        height=5cm,
        grid=major,
        xmajorgrids=false,
        ymajorgrids=true,
        major grid style={line width=0.2pt, draw=gray!20},
        symbolic x coords={GraspNet-1B, GSNet, HGGD, HGGD-MCU},
        xtick=data,
        ymin=0, ymax=100,
        ybar=1pt,
        ylabel={\tiny Average Precision},
        xlabel={},
        tick label style={font=\tiny},
        ylabel near ticks,
        legend pos=north east,
        legend style={font=\tiny},
        x tick label style={rotate=0, anchor=center}
    ]
        \addplot[black!99, pattern=dots, single ybar legend] coordinates {(GraspNet-1B, 10.55) (GSNet, 24.31) (HGGD, 24.59) (HGGD-MCU, 21.87)};
        \addplot[black!99, pattern=grid,single ybar legend,] coordinates {(GraspNet-1B, 11.15) (GSNet, 30.52) (HGGD, 30.46) (HGGD-MCU, 25.37)};
        \addplot[black!99, pattern=crosshatch,single ybar legend,] coordinates {(GraspNet-1B, 3.98) (GSNet, 14.23) (HGGD, 15.58) (HGGD-MCU, 6.51)};
        \legend{AP$_1$, AP$_{0.8}$, AP$_{0.4}$}
    \end{axis}
\end{tikzpicture}
		\caption{Measured AP for the ``Novel'' dataset.}
		 \label{fig:APnovel}
	\end{figure}
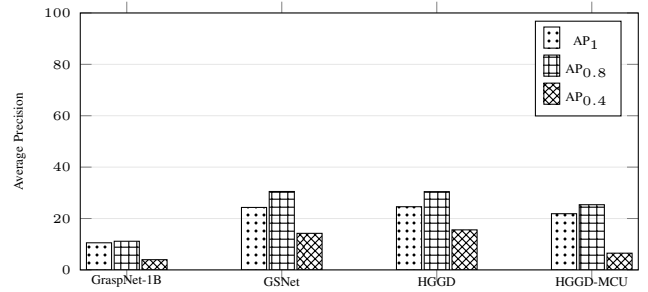

Figures \ref{fig:APseen}, \ref{fig:APsim} and \ref{fig:APnovel} demonstrate the recorded $AP_{\mu}$ where $\mu$ is the friction coefficient for the Seen, Similar, and Novel datasets respectively. To complement these results, Table \ref{tab:comparison} provides a broader comparison against state-of-the-art models. ``HGGD'' refers to the original HGGD implementation, while ``HGGD-MCU'' represents our reduced pipeline. 


\begin{figure}[t]
	   \centering
		\begin{tikzpicture}[scale=1]
        \begin{axis}[
        ybar,
        bar width=9.5pt,
        width=9cm,
        height=5cm,
        grid=major,
        xmajorgrids=false,
        ymajorgrids=true,
        major grid style={line width=0.2pt, draw=gray!20},
        symbolic x coords={GraspNet-1B, GSNet, HGGD, HGGD-MCU},
        xtick=data,
        ymin=0, ymax=100,
        ybar=1pt,
        ylabel={\tiny Average Precision},
        xlabel={},
        tick label style={font=\tiny},
        ylabel near ticks,
        legend pos=north east,
        legend style={font=\tiny},
        x tick label style={rotate=0, anchor=center}
    ]
        \addplot[black!99, pattern=dots, single ybar legend] coordinates {(GraspNet-1B, 26.11) (GSNet, 54.81) (HGGD, 53.59) (HGGD-MCU, 41.36)};
        \addplot[black!99, pattern=grid,single ybar legend,] coordinates {(GraspNet-1B, 34.18) (GSNet, 66.72) (HGGD, 64.12) (HGGD-MCU, 53.1)};
        \addplot[black!99, pattern=crosshatch,single ybar legend,] coordinates {(GraspNet-1B, 14.23) (GSNet, 46.17) (HGGD, 45.91) (HGGD-MCU, 18.73)};
       \legend{AP$_1$, AP$_{0.8}$, AP$_{0.4}$}
    \end{axis}
\end{tikzpicture}
		\caption{Measured AP for the ``Similar'' dataset.}
		\label{fig:APsim}
	\end{figure}
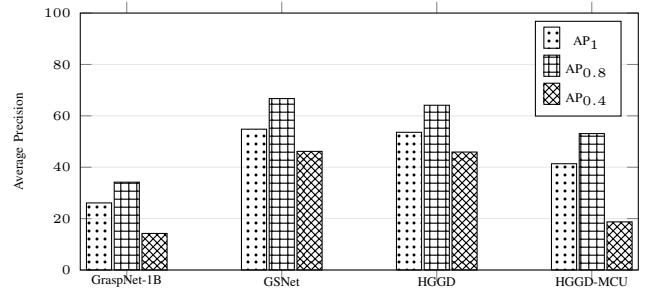


\subsection{Memory requirements}
\begin{figure}[h!]
    \centering
\begin{tikzpicture}
    \begin{axis}[
        width  = 0.45*\textwidth,
        height = 5cm,
        major x tick style = transparent,
        ybar=1pt,
        bar width=12pt,
        ymajorgrids = true,
        xlabel = {\tiny Compressed Models},
        ylabel = {\tiny L2 requirements (Bytes)},
        symbolic x coords={Original, Optimised},
        xtick = data,
        tick label style={font=\tiny},
        scaled y ticks = false,
        enlarge x limits=0.30, 
        ymax=2000000,
        legend pos=north east,
        legend style={font=\tiny,line width=.3pt,mark size=.3pt, row sep=-2pt},
        legend image code/.code={\draw[#1] (0cm,-0.1cm) rectangle (0.3cm,0.1cm);},
        ymode=log,
        minor y tick num=3, 
    ]
        \addplot[black!99, pattern=dots]
            coordinates { (Original, 864316.0) (Optimised, 864756.0)};

         \addplot[black!99, pattern=grid]
            coordinates { (Original, 1300000.0) (Optimised, 963092.0)};

         \addplot[black!99, pattern=crosshatch]
            coordinates {(Original, 1241932.0) (Optimised, 870840.0)};

         \addplot[black!99, style={black,fill=black,mark=none}]
            coordinates {(Original, 1280400.0)};

        \legend{LocalNet-MCU, ResNet-MCU, AnchorNet-MCU, PointNet-MCU}
    \end{axis}
\end{tikzpicture}
    \caption{L2 requirements for the GAP9}
    \label{fig:L2GAP9}
\end{figure}
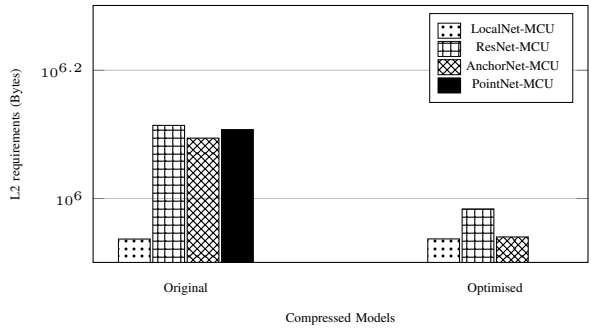

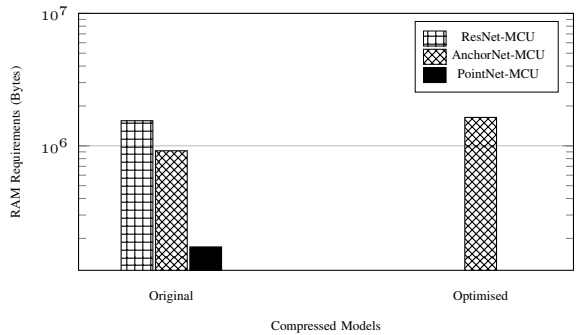
\begin{figure}[h!]
    \centering
\begin{tikzpicture}
    \begin{axis}[
        width  = 0.45*\textwidth,
        height = 5cm,
        major x tick style = transparent,
        ybar=1pt,
        bar width=12pt,
        ymajorgrids = true,
        xlabel = {\tiny Compressed Models},
        ylabel = {\tiny RAM Requirements (Bytes)},
        symbolic x coords={Original, Optimised},
        xtick={Original, Optimised},
        tick label style={font=\tiny},
        scaled y ticks = false,
        enlarge x limits=0.30, 
        ymax=10000000,
        legend pos=north east,
        legend style={font=\tiny,line width=.3pt,mark size=.3pt, row sep=-2pt},
        legend image code/.code={\draw[#1] (0cm,-0.1cm) rectangle (0.3cm,0.1cm);},
        ymode=log,
        minor y tick num=3, 
    ]


         \addplot[black!99, pattern=grid]
            coordinates { (Original, 1546272.0) (Optimised, 0)};

         \addplot[black!99, pattern=crosshatch]
            coordinates { (Original, 917504.0) (Optimised, 1638400.0)};

         \addplot[black!99, style={black,fill=black,mark=none}]
            coordinates {(Original, 172328.0) (Optimised, 0)};

        \legend{ResNet-MCU, AnchorNet-MCU, PointNet-MCU}

    \end{axis}
\end{tikzpicture}
    \caption{RAM requirements for the GAP9}
    \label{fig:RAMrewGAP9}
\end{figure}

Plots \ref{fig:FlashGAP9}-\ref{fig:RAMrewGAP9} use a logarithmic scale to highlight significant memory variations across models. The available memory capacity is indicated by the upper bound of the y-axis. 

On the GAP9 platform, flash memory usage (Figure \ref{fig:FlashGAP9}) is significantly reduced by partitioning the original models into smaller sub-models. All model variants fit well within the available flash capacity, indicating that flash is not a limiting factor in this setup.

Regarding L$2$ and RAM requirements (Figures \ref{fig:L2GAP9} and \ref{fig:RAMrewGAP9}), models exceeding the L$2$ cache limit are automatically mapped to RAM, ensuring execution remains within the overall memory constraints. Segmenting the original architectures greatly reduces both L$2$ cache and RAM usage.

\begin{figure}[h!]
    \centering
\begin{tikzpicture}
    \begin{axis}[
        width  = 0.45*\textwidth,
        height = 5cm,
        major x tick style = transparent,
        ybar=1pt,
        bar width=12pt,
        ymajorgrids = true,
        xlabel = {\tiny Compressed Models},
        ylabel = {\tiny Flash Requirements (Bytes)},
        symbolic x coords={Original, Optimised},
        xtick = data,
        tick label style={font=\tiny},
        scaled y ticks = false,
        enlarge x limits=0.30, 
        ymax=2000000,
        legend pos=north east,
        legend style={font=\tiny,line width=.3pt,mark size=.3pt, row sep=-2pt},
        legend image code/.code={\draw[#1] (0cm,-0.1cm) rectangle (0.3cm,0.1cm);},
        ymode=log,
        minor y tick num=3, 
    ]
        \addplot[black!99, pattern=dots]
            coordinates { (Original, 862232.0) (Optimised, 862672.0)};

         \addplot[black!99, pattern=grid]
            coordinates { (Original, 1343104.0) (Optimised, 707092.0)};

         \addplot[black!99, pattern=crosshatch]
            coordinates {(Original, 961868.0) (Optimised, 287160.0)};

         \addplot[black!99, style={black,fill=black,mark=none}]
            coordinates {(Original, 960144.0)};

        \legend{LocalNet-MCU, ResNet-MCU, AnchorNet-MCU, PointNet-MCU}
    \end{axis}
\end{tikzpicture}
    \caption{Flash requirements for the GAP9}
    \label{fig:FlashGAP9}
\end{figure}
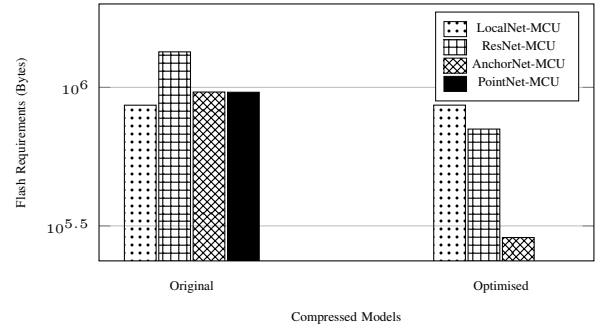

\subsection{Measuring inference time}
The GAP9 successfully processed the input of the HGGD-MCU model, achieving an average throughput of $740.47 \pm 0.0046$ milliseconds. Performance analysis revealed that PointNet-MCU was a significant bottleneck, as the original code is hardcoded to operate with a batch size of $1$, and modifying this results in errors. As a workaround, we performed $64$ passes through the network, which significantly impacted processing speed. Table \ref{tab:performance_metricsGAp} presents the inference times for each sub-model when executed on the GAP9.

\begin{table}[h]
    \centering
    \begin{tabular}{|>{\centering\arraybackslash}m{3cm}|c|}
        \hline
        \textbf{Model}& \textbf{Inference time (ms)} \\ \hline
        ResNet-MCU & 15.13\\ \hline
        AnchorNet-MCU& 44.96\\ \hline
        PointNet-MCU & 674.79\\ \hline
        LocalNet-MCU& 5.58\\ \hline
    \end{tabular}
    \caption{Inference time for each model in milliseconds}
    \label{tab:performance_metricsGAp}
\end{table}

\section{Discussion} \label{sec:discussion}

\subsection{Average Precision}
Validation on the GraspNet-1Billion dataset confirmed that our reduced pipeline, HGGD-MCU delivered competitive results. Despite a lower input data resolution, the model performed closely to state-of-the-art approaches, such as GSNet and the original HGGD, when $\mu = 1$ or $0.8$, with a noticeable performance drop at $\mu = 0.4$.

As listed in Table~\ref{tab:comparison}, methods such as \cite{EconomicGrasp} and \cite{wang2024} report higher accuracy scores. However, it is important to note prior works were not designed with resource-constrained hardware in mind. In contrast, our work explicitly targets deployment on low-power MCUs. While this makes direct comparison somewhat imbalanced, we include these baselines to position our approach within the broader context of grasp detection. The results highlight the practical trade-offs between model complexity and deployability on edge devices.


\subsection{Memory requirements}
The observed flash memory trends align with expectations. Techniques that preserve parameter count, such as lowering the resolution of the input data, maintain similar flash usage.
In contrast, quantisation and model segmentation significantly reduce the number of parameters, leading to a substantial reduction in flash utilisation.

RAM usage is primarily driven by intermediate activations during inference. Reducing network depth does not necessarily lower RAM demands, especially when memory usage is dominated by feature maps.
However, techniques like resolution reduction and quantisation directly reduced feature map size or precision, thereby reducing RAM consumption. Model partitioning also reduces peak memory usage, since each sub-model is executed independently.

Hierarchical memory architectures further improve efficiency. When sub-models exceed the on-chip L2 cache, they spill into system RAM, ensuring that they remain fully operable. At the same time, smaller models that fully fit within L2 benefit from lower latency due to faster memory access. These findings confirm that careful selection and tuning of optimisation strategies are critical for efficient memory use on resource-constrained platforms such as the GAP9.

\begin{table*}[h]
    \centering
    \begin{tabular}{|>{\centering\arraybackslash}m{1.5cm}|c|c|c|c|c|c|c|c|c|}
        \hline
         & \multicolumn{3}{c|}{\textbf{Seen}} & \multicolumn{3}{c|}{\textbf{Similar}}  & \multicolumn{3}{c|}{\textbf{Novel}}\\ \hline
    	\textbf{Method} & \textbf{AP$_{1}$} & \textbf{AP$_{0.8}$} & \textbf{AP$_{0.4}$} & \textbf{AP$_{1}$} &  \textbf{AP$_{0.8}$} & \textbf{AP$_{0.4}$} 
        & \textbf{AP$_{1}$} & \textbf{AP$_{0.8}$} & \textbf{AP$_{0.4}$} \\ \hline
\cite{chen2023efficient} & $64.45$ &  $72.81$ & $61.16$ & $53.59$ & $64.12$ & $45.91$ & $24.59$ & $30.46$ & $15.58$ \\ \hline
\cite{wang2021graspness} & $67.12$ & $78.46$ & $60.9$ & $54.81$ & $66.72$ & $46.17$ & $24.31$ & $30.52$ & $14.23$ \\ \hline
\cite{EconomicGrasp} & $68.21$ & $79.6$ & $63.54$ & $61.19$ & $73.60$ & $53.77$ & $25.48$ & $31.46$ & $13.85$ \\ \hline
\cite{wang2024} & $74.33$ & $85.77$ & $63.89$ & $64.36$ & $76.76$ & $55.25$  & $27.56$ & $34.09$ & $20.23$ \\ \hline
\textbf{This work} & $42.19$ & $51.94$ & $22.98$ & $41.36$ & $53.1$ & $18.73$ & $21.87$ & $25.37$ & $6.51$ \\ \hline
    \end{tabular}
    \caption{AP Comparison with State-of-the-Art Models}
    \label{tab:comparison}
\end{table*}

\subsection{Inference time}
Among the validated models, PointNet-MCU was the most computationally expensive, accounting for approximately $90\%$ of the total inference time. This cost is attributed to the hard-coded batch size of $1$, as NNTool failed when attempting to set a batch size of $64$. Future updates to the tool may address this issue, while alternative workarounds could be explored.


With an inference time of $740$ms, the model demonstrates speeds that can be considered adequate for many grasping tasks, particularly in scenarios where robotic arms operate at slower speeds. For instance, in tasks involving delicate handling or precise placements, the robot's movement often dictates the overall operation speed rather than the computational latency. Further optimisations, such as reducing the computational overhead of PointNet or parallelising certain operations, could reduce inference times even further, making the system more versatile and suitable for real-time applications. This study underscores the potential of the GAP9 as a reliable and efficient platform for edge AI tasks in robotics.

\subsection{Exploration}
We applied a range of optimisation techniques to reduce both inference time and memory usage. Despite these efforts, certain layers proved incompatible with the deployment toolkit, and some model variants exceeded the GAP9’s memory capacity or relied on unsupported operations.

To address near real-time performance requirements, we partitioned the grasp perception pipeline into four sub-models, each responsible for a specific stage of the original architecture. This approach allowed a pipelined inference but introduced overhead due to the transfer of intermediate results between stages. In principle, this latency could be mitigated through the deployment of multiple GAP9s in parallel, each running a pipeline stage. 
Our findings suggest that model decomposition and offloading to a host processor or multiple MCUs is a promising direction for increasing throughput. 



\section{Conclusions} \label{sec:conclusion}
This study explored the feasibility of deploying a state-of-the-art robotic grasping perception model on resource-constrained MCUs. Our work focused on optimising the HGGD model for 6D grasping tasks on the GAP9 SoC.

We demonstrated that by reducing the input size and partitioning the model we achieved a strong performance on the GraspNet-1Billion dataset. 

Reducing the input resolution lowered significantly the model's memory requirements with only a modest impact on performance.
Model partitioning further reduced both flash and memory usage at no additional performance cost. However, this came at the cost of increased inference time, as models have to be loaded and unloaded from the MCU as the model progresses. This limitation could be mitigated by parallelising execution across multiple MCUs. 
Additionally, quantisation reduced significantly flash and memory footprints.


Future work will focus on extending these findings to a broader range of MCUs and evaluate system performance in robotic hardware setups. We also encourage further investigation into structural optimisations, such as layer pruning, which may yield additional gains at the cost of retraining. These directions will help advance the deployment of complex vision-based robotic systems on low-power, embedded hardware.

\bibliographystyle{unsrt}

\end{document}